\documentclass[12pt]{article}
\usepackage[margin=1in]{geometry}

\usepackage{
	amsmath,
	amsthm,
	amssymb,
	mathrsfs,
	MnSymbol,
	empheq,
	graphics,
	graphicx,
	enumitem,
	bbm,
	xcolor,
	tikz,
	relsize,
	float,
	verbatim,
	hyperref,
	color,
	courier,
	listings,
	datetime,
	lastpage,
	titling,
	caption,
	subcaption,
	booktabs,
        natbib}

\usepackage[english]{babel}
\usepackage[utf8]{inputenc}
\usepackage[T1]{fontenc}
\usepackage[framemethod=default]{mdframed}
\usepackage[super]{nth}
\usepackage[parfill]{parskip}

\hypersetup{
	colorlinks,
	citecolor=black,
	filecolor=black,
	linkcolor=black,
	urlcolor=blue
}

\newcommand{\N}{\mathbb{N}}

\newcommand{\R}{\mathbb{R}}

\newcommand{\ba}{\[\begin{aligned}}
\newcommand{\ea}{\end{aligned}\]}

\newcommand{\xvec}{\mathbf{x}}
\newcommand{\uvec}{\mathbf{u}}

\newcommand{\piex}{\pi^\mathrm{ex}}
\newcommand{\pilt}{\pi^\mathrm{LT}}

\title{Multi-market Energy Optimization with Renewables via Reinforcement Learning} 
\author{Lucien Werner, Peeyush Kumar}

\begin{document}

\maketitle


\section{Introduction}
The decarbonziation of the bulk electric power system, whose generation mix still mainly consists of gas and coal-fired thermal generation, is mainly being driven by the rapidly falling costs of utility-scale wind and solar. These variable generation sources produce no emissions when producing energy but unlike conventional thermal power plants, they are not controllable---also called dispatchable---and therefore can not modulate their output at will. In some sense, nature determines the supply trajectory over time. 

At low penetrations (e.g., 5-20\%), variable renewable energy (VRE) can be accommodated by the existing grid. However at the high penetrations needed for a low-carbon grid, the variability and unpredictability of solar and wind pose major challenges to a \textit{reliable} power supply that is able to instantaneously meet electrical demand. Even momentary supply-demand imbalances in electrical grids can cause cascading blackouts and thus reliability is the major challenge for a future grid powered mainly by renewables.

Energy storage in the form of batteries or pumped hydro has been widely studied as a well to smooth out the fluctuations in VRE production. Co-locating energy storage with renewable generation is a way for renewable generation to make firm commitments in energy markets. This is important as it reduces the amount of uncertainty and variability that the renewable generator introduces into the balance of the power grid. 

The operators of combined renewable plus storage plants seek to optimize the revenue they obtain from energy markets. This will often include long-term contracts with penalties for non-delivery (like PPAs), day-ahead and real-time markets, as well as fast-timescale markets for frequency regulation. In this work we study the combination of long-term and real-time markets, although the modeling and optimization framework can be extended to include additional or faster-timescale markets.

\subsection{Contributions}
This work contributes to two principle areas: energy market optimization and reinforcement learning. The main novelty of the work is a stochastic optimization framework for performing multi-market energy optimization. This is the task of computing optimal energy dispatches in several, coupled energy markets under uncertainty. Examples of such markets are long-term forward contracts like power purchase agreements (PPAs) and capacity markets, as well as more real-time scheduling markets like the day-ahead, real-time, and frequency regulation markets. Each of these markets has a distinct transaction timescales, which ranges from decades to seconds. The uncertainty arises from the stochasticity of energy production by renewables (ultimately caused by weather) and the unpredictability of short-timescale market prices.

Applying reinforcement learning to solve stochastic energy optimization problems in an area of growing interest in the research community. However, several challenges remain: specifically, learning \textit{safe} policies that respect physical and hard constraints and managing variability in exogenous information during policy learning. In order to derive \textit{practical} control algorithms with an RL approach, both of these issues must be surmounted.

The challenges motivate the following research agenda:

\begin{itemize}
    \item Form multi-objective stochastic optimization framework for modeling multi-market energy scheduling problem;
    \item Guarantee hard constraint satisfaction for power systems;
    \item Handle variability in exogenous data for prices and renewable generation.
\end{itemize}

\subsection{Related Work}
Reinforcement learning has increasingly been applied to complex control and sequential decision-making problems in uncertain environments \cite{arulkumaran2017deep}. Researchers have also explored applying RL in the power systems context for problems such as voltage control, frequency control, and energy scheduling. \cite{zhang2019deep, yang2020reinforcement}. It is widely recognized that safety and scalability and key issues that arise for algorithms for power system applications \cite{chen2021reinforcement}. In general, RL policies are neither guaranteed to satisfy real-world system constraints during training nor during run-time. Our work directly addresses this challenge. 

Existing works that focus on deep reinforcement learning for energy storage management include \cite{shang_stochastic_2020, wang_energy_2018, oh_reinforcement-learning-based_2020, cao_deep_2020, yang_deep_2020, mbuwir2017battery}. None of these approaches considers a general modeling framework that can accommodate both storage and renewables for arbitrary systems. 

Recent research on constraint satisfaction for power systems has two notable contributions: \cite{chen2021enforcing} and \cite{zheng2021safe}. The former uses differentiable convex layers to project actions onto a feasible set as part of the policy learning. The latter takes a similar approach but proposes a novel projection layer in the neural network that does not require computing the solution map of a large convex optimization problem. The approach we take in this paper is complementary to these proposals and exploits similar properties of the projection problem (such as control-affine-ness) to incorporate the projection step into the system dynamics. 

Finally, \cite{powell2014clearing, powell2015tutorial, powell2021reinforcement} form a body of literature that focuses on multi-objective stochastic optimization for multiple timescales. The solution approaches utilize classical dynamic programming or approximate dynamic programming methods which either not suitable for the continuous state-action space environment of our problem or uses discretization schemes that are intractably large.  

\section{A Primer on Electricity Markets}
Electricity markets are complex and varied, with regional differences in terms of regulatory structures and market design but they share some common similarities. For instance in the United States at a high level, these markets can be divided into two primary types: regulated and deregulated markets \cite{litvinov2019electricity}.

In traditionally regulated markets, utilities are vertically integrated and operate as monopolies within their territories, meaning that they own the electricity generators and power lines, and customers only have the option to buy power from them. To prevent overcharging, state regulators oversee how these utilities set electricity prices, which are determined based on the utility's operating and investment costs as well as a "fair" rate of return on those investments. These rates must be approved by the state's public utilities commission. Utilities in these regions must also seek state approval for power plant investments, a process that often requires demonstrating the necessity of proposed investments through an integrated resource planning (IRP) process. This approach to regulation and monopolization means that customers bear the risk of investments as utilities can recover their costs through rates, regardless of how power plants perform.

Even though these vertically integrated utilities generate their own electricity, they can trade with other utilities during times of need, especially when it's cheaper to purchase power than to generate it using their own facilities. This is common in the western and southeastern United States, and such transactions are regulated by the Federal Energy Regulatory Commission (FERC).

Deregulated markets, on the other hand, began emerging in the 1990s when many US states decided to create competition and lower costs \cite{hogan2016virtual}. This restructuring required electric utilities to sell their generating assets, leading to the creation of independent energy suppliers. The utilities retained their transmission and distribution assets, which remain regulated In these areas, retail deregulation allows customers to select their own electric supplier, introducing competition for retail electricity prices. However, transmission and distribution services are still provided by the local utility company. Contracts with independent suppliers can have both pros and cons for consumers, and only the generation portion of a customer's utility bill is set competitively.

In contrast to regulated states that plan for investment, deregulated states use markets to determine which power plants are necessary for electricity generation. Wholesale deregulation led to the creation of centralized wholesale \cite{cervigni2013wholesale} markets in which power is sold by generators and bought by entities that sell it to consumers. Investment risks in power plants fall to the electric suppliers and not to customers in deregulated markets.

Following deregulation, Regional Transmission Organizations (RTOs) replaced utilities as grid operators and became the operators of wholesale markets for electricity. Many of these RTOs operate markets that span multiple states and are therefore regulated by FERC, with the exception of the Texas RTO, the Electric Reliability Council of Texas (ERCOT).

In deregulated retail utilities, electricity is purchased at market-determined wholesale prices and sold to customers at similarly determined retail prices. RTOs typically run three kinds of markets that determine these wholesale prices: energy markets, capacity markets, and ancillary services markets.

\textbf{Energy Markets}: The main product in these markets is electricity itself. Generators offer their available capacity for each hour of the day, while utilities and other entities bid on the capacity they need to serve their customers. The market operator, typically a Regional Transmission Organization (RTO), then matches supply with demand and determines the price for each hour.\\
\textbf{Capacity Markets}: These markets operate in parallel to energy markets and involve the trade of generation capacity, rather than energy itself. The purpose of capacity markets is to ensure that there is enough power generation capacity available to meet peak demand. They provide a means for generators to be compensated for their availability, regardless of whether they are actually generating electricity. The rules for capacity markets vary by region, but they generally involve auctions in which generators bid their available capacity, and utilities and other entities pay for the capacity they need to ensure reliability \cite{fabra2018primer}.
\\\textbf{Ancillary Services Markets}: These markets involve the trade of services that are essential to maintaining the reliability and security of the power grid, but that are not captured in the energy or capacity markets. Examples of ancillary services include frequency regulation, voltage control, and operating reserves. Like energy and capacity, ancillary services are typically coordinated by RTOs, and prices are determined based on supply and demand \cite{helman2009rtos}.

    \begin{figure}[h]
        \centering
        \includegraphics[width=0.7\columnwidth]{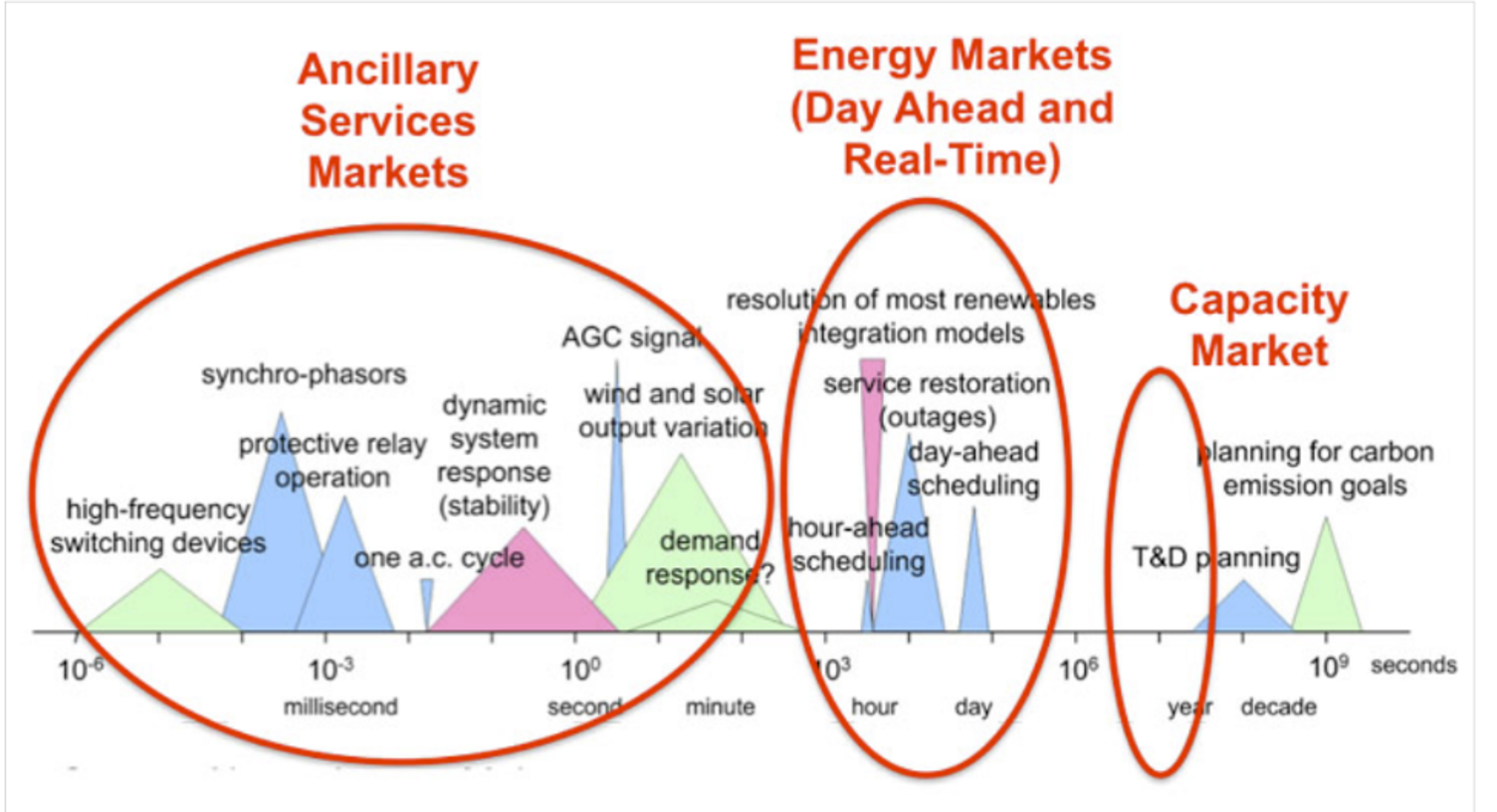}
        \caption{Electricity Markets}
        \label{fig:emarkets}
    \end{figure}

Electricity markets operate on various timescales, with day-ahead markets facilitating transactions based on the next day's forecasted load, while real-time markets adjust to meet the actual demand for electricity, which can vary minute to minute. Capacity markets, plan for peak electricity demand years in advance, often through an auction system \cite{tang2022multi}. Figure~\ref{fig:emarkets} display the multiple timescales of these electricity markets. 

\section{Problem Formulation}
In this section we cast the multi-market energy optimization problem as a stochastic optimization problem \textit{over policies}. In order to accommodate the uncertainty in the exogenous prices and VRE production, we consider searching for an optimal policy (of system state and information) rather than seeking an optimal sequence of decisions, as is done in deterministic optimization. 

First we pose the problem as a Markov decision process (MDP) and detail the models for the energy system and the environment (including exogenous processes). In Section \ref{sec:rl_alg} we present the method for solving the stochastic optimization problem derived from the MDP. 

We consider a discrete, finite time horizon of length $T$ indexed by $t$:
\[
t \in \{1,\ldots,T\}.
\]
We assume that the state of the system and the information state (exogenous and endogenous) are unchanged over the duration of interval $t$. The length of interval $t$ is determined by the granularity of the input data; for example this could be 5 mins if nodal price data is used or 1 hr if exchange market data is used. 

The system is modeled as a microgrid attached to the bulk power grid at a single bus. This microgrid includes renewable generation, energy storage, and conventional controllable loads or generation. This framework could describe a variety of realistic situations, including a hybrid renewable-storage power plant, a commercial building with rooftop solar and aggregated controllable electric vehicles, or an islandable microgrid. 

Following the notation from  \citet{powell2015tutorial} we represent the controllable state of the system---which includes the battery state-of-charge (SOC), the renewable energy injections, the injections from controllable loads/generation, and the commitments to both markets--with the vector $X_t$\footnote{In general $X_t$ is a function.}. The exogenous variables are real-time prices and maximum renewable energy production and are collected in the random vector $W_t$. The available actions are given by the vector function $U_t$; actions include the injections (charge/discharge) from storage. Finally the system transition function, which is deterministic given realized uncertainty $W_t$, is given by
\[
X_{t+1} = f(X_t,U_t,W_t).
\]
We will detail all of these components subsequently in this section.

\subsection{Environment Model}\label{sec:env_model}
A key feature of our model is state and action constraints. Such constraints are crucial in the context of power systems, as there are hard physical constraints like max charge/discharge for energy storage that cannot be violated as well as physical laws like conservation of power (at the PCC) that---if violated---would render a policy meaningless. 

Now we present the notation and constraints for all of the system components.


\subsubsection{Energy storage}
There $N_B$ storage devices, indexed with $b \in \{1,\ldots,N_B\}$.\footnote{We will use the terms storage and battery interchangeably, even though the storage device will be modeled to allow for more diverse storage technologies such as pumped hydro, compressed gas, and mechanical storage.} The state-of-charge (SOC) of each storage device, which represents how much energy in Watt-hours (Wh) the device holds at time $t$, is denoted $X_{b,t} \in \R_+$. The vector of SOC values over the entire time horizon is denoted $\xvec_b$.\footnote{The boldface notation generally represents vectors of \textit{decisions}. Although the state is in general a function we will sometime use the vector notation for convenience.} The SOC of each storage resource is assumed to be bounded in each time period between positive lower/upper bounds:
\begin{equation}\label{eqn:SOC_limits}
    \underline{x}_{b,t} \leq X_{b,t} \leq \overline{x}_{b,t}
\end{equation}

Note that when $\underline{x}_{b,t} = \overline{x}_{b,t}$ for some $t$, the SOC is fixed. This can be used to enforce initial or terminal SOC constraints. 

Charge/discharge actions change the SOC of the storage device. We model charge and discharge actions separately, using $U^+_{b,t} \in \R_+$ for the former and $U^-_{b,t} \in \R_+$ for the latter. Both charge and discharge values are restricted to be positive. Vectors of charge/discharge decisions over the entire time horizon are given by $\uvec_b^+$ and $\uvec_b^-$. Charging limits are assumed in each time period:
\begin{align}\label{eqn:cd_limits}
    \underline{u}^+_{b,t} \leq u^+_{b,t} &\leq \overline{u}^+_{b,t}\\
    \underline{u}^-_{b,t} \leq u^-_{b,t} &\leq \overline{u}^-_{b,t}
\end{align}

The SOC and charge/discharge actions are related through the following SOC update equation from time $t$ to time $t+1$:
\begin{equation}\label{eqn:SOC_update}
x_{b,t+1} = x_b + f_b^+(u^+_{b,t}) - f_b^-(u^-_{b,t})
\end{equation}
The functions $f^+$ and $f^-$ map to positive values and represent the possibly non-linear dependence of the SOC on the charge/discharge actions. A simple version of these functions is
\begin{align*}
    f_b^+(u^+_{b,t}) &= \eta^+_b u^+_{b,t}\\
    f_b^-(u^-_{b,t}) &= \eta^-_b u^-_{b,t}
\end{align*}
where $\eta_b^\pm$ are constant charge/discharge efficiencies. In general, $f^\pm$ can be non-linear functions of both $u$ and $x$. Also notice that (\ref{eqn:SOC_update}) does not preclude simultaneous charging and discharging. This may be desirable or even optimal in some context (e.g., for pumped hydro storage) but can be undesirable for chemical storage. \citet{nazir2021guaranteeing}, \citet{garifi2020convex} discuss conditions under which simultaneous charging and discharging can occur.

We also model a cost of charge/discharge for each energy storage resource $C_b$. In general, the cost function can depend explicitely on the SOC and the particular charge/discharge actions, as well as any of the previously realized actions. Therefore we will represent the cost function with the notation
\[
C_{b,t}(\xvec_b[0:t], \uvec_b^+[0:t], \uvec_b^-[0:t])
\]
where $\xvec_b[0:t]$ is the sub-vector of SOC values up to time $t$. 

The difficulty in modeling storage devices comes from the forms of $C_{b,t}$ and $f_b^{\pm}$. These functions are non-linear and in the case of the cost function, depends on the entire charging profile rather than just a discrete charging action. Our model allows either the typical linear simplifications or black-box simulators that ingest inputs and return a cost. In Section \ref{section:experiments} we discuss the particular choices we make in the construction of our simulator.

\subsubsection{Renewable Generation}
There are $N_R$ renewable generators, indexed with $r \in \{1,\ldots,\N_R\}$. The variable $x_{r,t} \in \R_+$ represents the power (W) that generator $r$ produces in time $t$. The vector of generation of the time horizon is denoted $\xvec_r$. The power from the generator can be curtailed but cannot be increased beyond the available power. The max available power in any period is upper bounded by the nameplate capacity. Therefore we impose the following constraints on $x_{r,t}$:
\[
0 \leq x_{r,t} \leq \overline{x}_{r,t} \leq x_r^\mathrm{nameplate}
\]

Given the fast ramping capabilities of renewable generation, no ramping constraints are placed on generation quantities between successive time periods. Further, we assume that there can be curtailment costs but no variable operation costs. 

\subsubsection{Controllable loads and generation}
There are $N_C$ controllable loads or generators, indexed with $i \in \{1,\ldots,N_R\}$. These are simply power injections (either positive or negative) that can be dispatched by the controller. The variable $x_{i,t} \in \R$ represent the power in watts that generator/load $i$ produces/consumes in time $t$. The vector of generation over the time horizion is denoted $\mathbf{x}_i$. The following constraints are imposed on $x_{i,t}$:
\[
\underline{x}_{i,t} \leq x_{i,t} \leq \overline{x}_{i,t}
\]
Fixed injections can be specified by setting $\underline{x}_{i,t} = \overline{x}_{i,t}$.

\subsubsection{System constraints}
In the system model, the microgrid is connected to the bulk power grid at a single point of common coupling (PCC). Therefore the total injection at this point must satisfy the laws of powerflow. In particular,
\[
x_t^\mathrm{PCC} = \sum_b (u_{b,t}^- - u_{b,t}^+)+ \sum_r x_{r,t} + \sum_i x_{i,t}
\]
By convention we assume that $x_t^\mathrm{PCC}
 > 0$ represents an outflow into the grid and $x_t^\mathrm{PCC}
 < 0$ represents an inflow. 
 
For the multi-market optimization, $x_t^\mathrm{PCC}$ is divided between real-time and long-term quantities. However, this is just an accounting issue. The producer will decide in each time what fraction of $x_t^\mathrm{PCC}$ to commit to each market. However, the purpose of the RL policy we seek to learn in this work is to optimize this tradeoff, and therefore the power balance equation is 
\[
x_t^\mathrm{LT} + x_t^\mathrm{RT} = \sum_b (u_{b,t}^- - u_{b,t}^+)+ \sum_r x_{r,t} + \sum_i x_{i,t}
\]
Note that microgrid network losses and congestion are not considered in this formulation, mainly to simplify the exposition. Conceptually these are simple to include in a linearized form, with a more involved linear equality constraint for power balance and linear inequality constraints for power flow constraints being added.  It is a reasonable assumption for a small microgrid whose devices are connected to a common busbar to be a copperplate (i.e., no congestion, no loss) network. 

\subsection{Uncertainty}
The principle challenge in solving the stochastic optimal control problem posed in the previous section is accommodating the uncertainty of the exogenous variables. There are two uncertain exogenous processes: the maximum available renewable generation $\overline{x}_{r,t} \ \forall r$ and the real-time price $\pi_t^\mathrm{RT}$. Neither of these random variables is stationary and their values at a given $t$ are dependent on some underlying non-stationary process. 

We discuss each source of uncertainty in greater detail next.

\subsubsection{Renewable generation}
\begin{figure}[h]
        \centering
        \includegraphics[width=0.8\columnwidth]{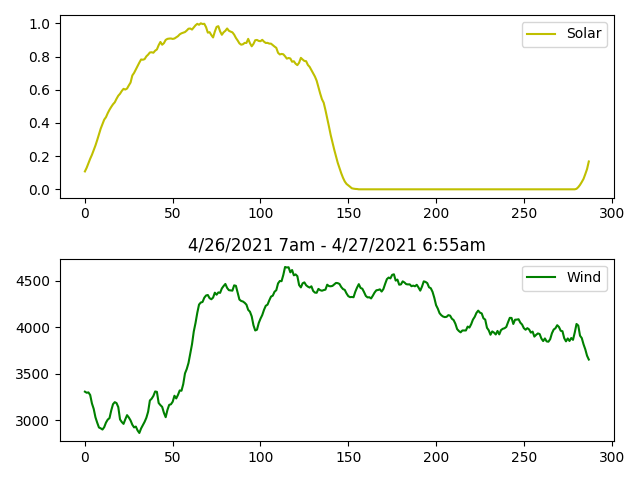}
        \caption{Sample solar and wind generation profiles. Solar is scaled between 0 and 1.}
        \label{fig:renewable_example}
    \end{figure}

The power produced by each renewable (e.g., solar or wind) generator in the microgrid is fundamentally uncertain. In the case of solar, the generation profile over the course of a day generally follows a nominal trajectory, peaking at midday and 0 at night, with fluctuations due to atmospheric conditions like clouds. Solar availability is also determined by the time of year, although the effect of the season on power output can be computed precisely. Thus, solar power can generally be forecast well with the exception of sometimes sharp dips in power due to local effects such as passing clouds or shadows. On the other hand, wind power is significantly less predictable and variable. Nonetheless wind can still be forecast in the short term with reasonable accuracy. For more details see \citep{kumar2021micro}. Figure \ref{fig:renewable_example} shows sample generation profiles for wind and solar at a location in California.
    
\subsubsection{Real-time prices}
The real-time price that the producer faces is generated by the system operator (e.g., ISO or RTO) every time the real-time market is cleared, which is typically 5 to 15 min intervals. This price is derived from a system-wide security constrained economic dispatch problem that can take into account system operating conditions like network congestion and losses as well as load and generation from all participants in the network. There is usually a price-taking assumption on producers, where they are assumed to not be able to manipulate the LMP. Although this assumption is not realistic in practice and in transmission constrained zones/nodes, individual generators may be able to increase the price through strategic withholding, \cite{navid2012market} we do not consider this possibility in our work. Strategic behavior on the part of generators is a broad topic that is the subject of much attention from the ISOs. Manipulative bidding is monitored by ISOs and there are typically penalties for participants that do so. 

\begin{figure}[h]
        \centering
        \includegraphics[width=0.8\columnwidth]{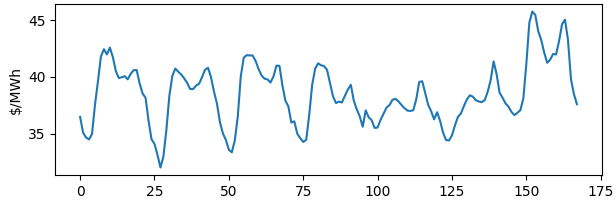}
        \caption{Nordpool hourly price over 1 week.}
        \label{fig:exchange_price}
    \end{figure}

We consider two types of uncertain price in the work. The first is an exchange price, which does not have a locational component, and can be found in European-style balancing markets, e.g., Nordpool. Exchange prices typically exhibit daily periodicity and track the demand in the market. An example is given in Figure \ref{fig:exchange_price}. 

\begin{figure}[h]
        \centering
        \includegraphics[width=0.8\columnwidth]{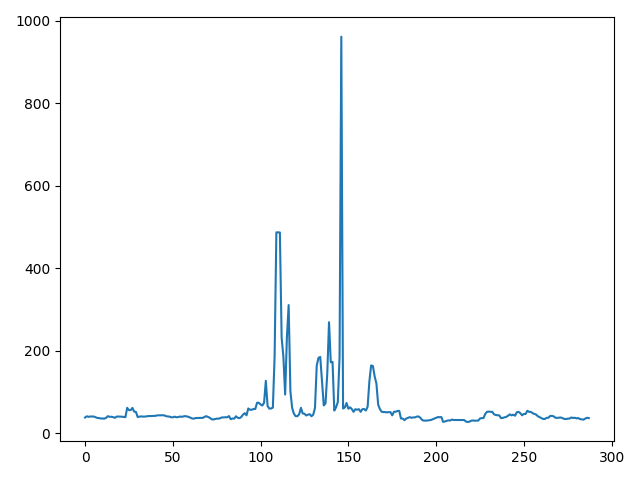}
        \caption{5-min LMP at a CAISO node over the course of one day. Price spikes up to the price cap of approx \$1000/MWh are likely due to temporary ramp-induced shortages.}
        \label{fig:LMP}
    \end{figure}

A second type of price is a locational marginal price (LMP), which includes congestion and loss components. The LMP is specific to a particular location (node) in the power network. By assumption, the producer in this work faces a single LMP as the PCC for the microgrid is connected to a single network bus. The variability in the LMP is significantly more challenging than that exhivited by the exchange price. This is due to binding congestion constraints that cause spikiness or discontinuity in the price. Further, LMPs do not enjoy the spatial averaging and smoothing present in exchange prices. An example of the LMP is given in Figure \ref{fig:LMP}

\subsection{Markov Decision Process}
We represent the evolution of the energy system through time as a Markov Decision Process (MDP). The MDP is the following 5-tuple: 
\[
(\mathcal{X}, \mathcal{U}, f, R, W_t)
\]

Consider $\mathcal{X}_t \subseteq \mathcal{X}$, $\mathcal{X}_t(W_t)$ is the time-varying state space, which depends deterministically on the realization of the exogenous variable. 
\[
X_t \in \mathcal{X}_t(W_t)
\]

Analogously we assume that there are possibly time-varying action constraints on $U_t$:
\[
U_t \in \mathcal{U}_t
\]
The transition dynamics are given by 
\begin{equation}\label{eqn:transition_dynamics}
    X_{t+1} = f(X_t, U_t, W_t)
\end{equation}
The reward function is the profit that the operator obtains in each time for delivering energy into the two markets:
\begin{equation}\label{eqn:reward}
    R_t(X_t,U_t) = P^\mathrm{RT}(X_t) + P^\mathrm{LT}(X_t^\mathrm{LT}) - C(X_t,U_t)
\end{equation}
The first two terms of $R_t$ are the profit/loss functions for the real-time and long term markets and the last term is the cost function, which has terms for storage degradation and renewable curtailment. The exact forms of these functions was discussed in Section \ref{sec:env_model}.

Note that the way we have defined the MDP explicitly depends on the exogenous uncertainty $W_t$. The state space, reward, and transition dynamics are deterministic functions given a realization of $W_t$. 



We assume that the randomness is from a distribution $\Omega$. In order to have a physically feasible system, the support of $\Omega$ is assumed to be finite. In addition, we have state and action constraints:

The reinforcement learning problem is to learn an (optimal) policy $\pi \in \Pi$ to determine the action $U_t$ in each time. That is,
\begin{align*}
\arg \max_{\pi \in \Pi} \quad &\sum_{t=0}^T\mathbb{E}_{W_t}\left[ R_t(X_t,U_t^\pi(X_t))\right] \\
\text{s.t.} \quad & X_{t+1} = f(X_t, U_t^\pi(X_t), W_t) \quad \forall t \\
&X_t \in \mathcal{X}_t(W_t) \quad \forall t \\
& U_t^\pi(X_t) \in \mathcal{U}_t \quad \forall t
\end{align*}
In the next section we discuss the algorithm we use to solve this stochastic optimization problem and find optimal $\pi^* \in \Pi$. 

\section{Reinforcement Learning Algorithm}\label{sec:rl_alg}
We first give some background on how the reinforcement learning algorithm interacts with the power system environment described in the previous section and then give details on the particular gradient-based policy method we exploit. 

    \begin{figure}[h]
        \centering
        \includegraphics[width=0.7\columnwidth]{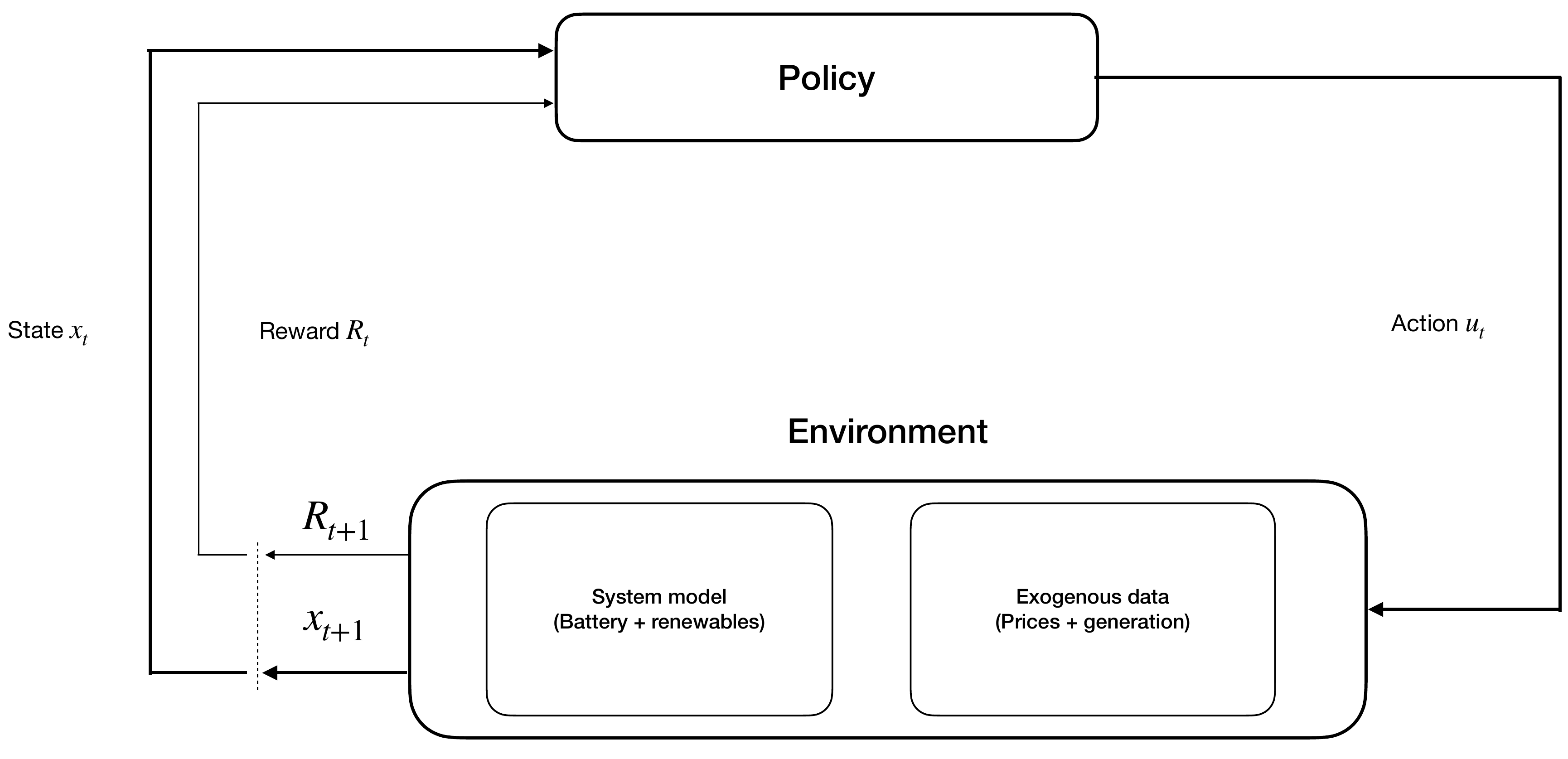}
        \caption{Interaction during learning between system environment and RL policy.}
        \label{fig:interaction}
    \end{figure}
    
The key aspect of our RL architecture is that the action $U_t$ returned by the policy is projected onto the feasible state-action space within the environment as part of the transition dynamics $f$. This is possible because the transition dynamics are assumed to be \textit{control-affine}; that is affine in the action $U_t$. 

\subsection{Proximal Policy Optimization (PPO)}

Proximal Policy Optimization (PPO) is a type of Reinforcement Learning algorithm that seeks to optimize policy functions in a more sample-efficient and stable way compared to other policy gradient methods. Traditional policy gradient methods make significant changes to the policy at each update, which can lead to unstable learning and poor performance. PPO addresses this by limiting the changes to the policy during each update, ensuring the new policy does not deviate significantly from the old policy. This is achieved using a specialized objective function that penalizes drastic changes. PPO, therefore, balances exploration of the environment and exploitation of the current policy, leading to reliable and efficient learning, which has made it a popular choice for a variety of complex tasks in diverse domains, see \cite{schulman2017proximal}. We will add more details about the specific adaptation of PPO for this case.

\section{Experiments}\label{section:experiments}
To make our problem setting accessible to a wide array of standard RL algorithms, we implement the energy environment described in this section as an OpenAI gym environment \cite{brockman2016openai}. 
\subsection{Case Study}
\begin{figure}[h]
	\centering
	\includegraphics[width=\textwidth]{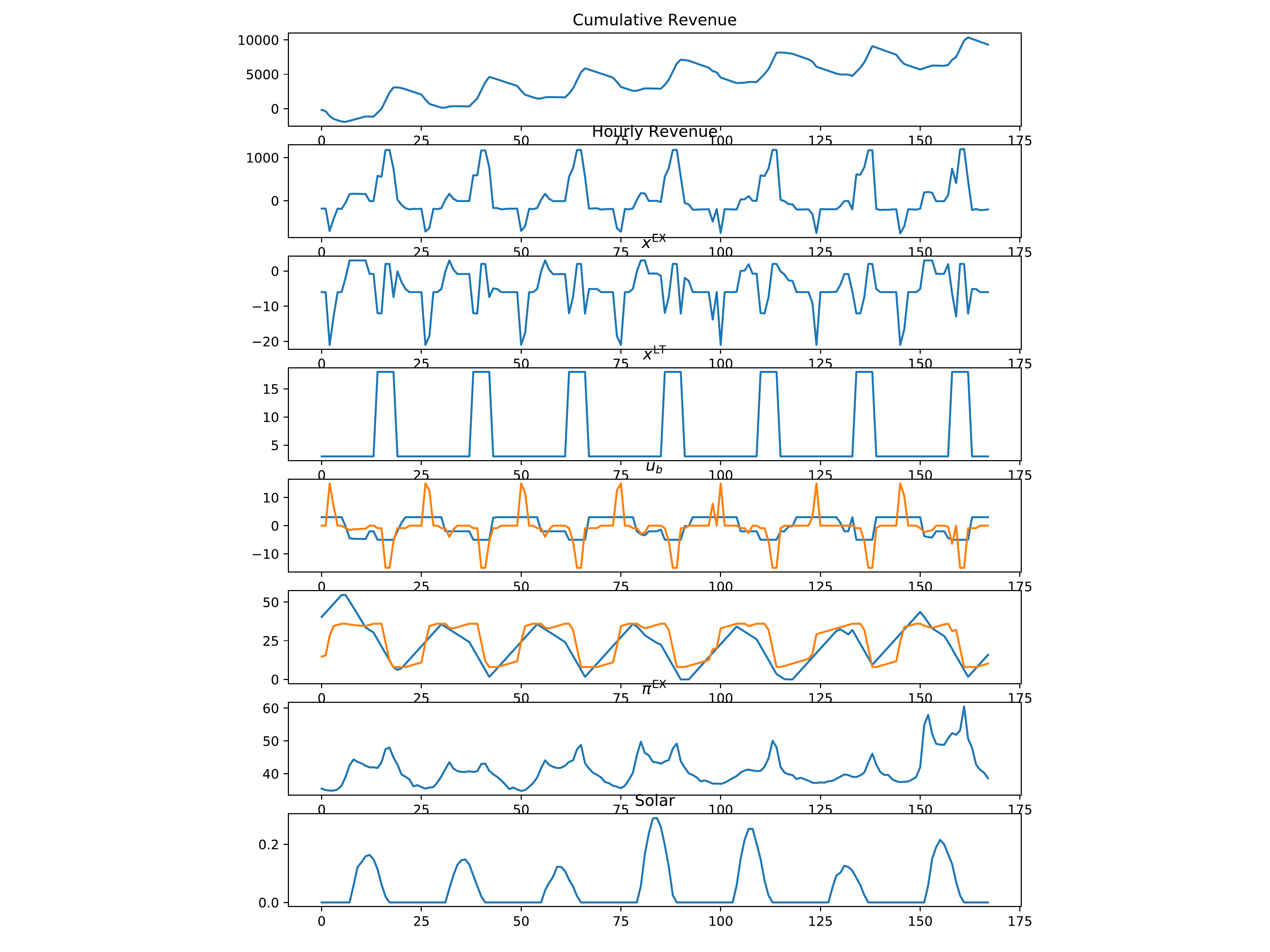}
	\caption{Time series of system evolution over 1 week at hourly intervals. The top two panes show the revenue the operator receives over time, the next two panes show the exchange and long-term market commitment quantities, respectively. The next two show the charge/discharge rates of the two batteries and their states-of-charge. Finally, the last two show the exogenous data; in particular exchange price fluctuations and the solar generation profiles.}
 \label{fig:results}
\end{figure}

We implement a small microgrid with a renewable generator (i.e., solar and wind) and two batteries: one with a large capacity but slow charge/discharge and one with a small capacity and fast charge/discharge. The results of the training are shown in Figure \ref{fig:results}.

\newpage
\section{Smoothing out the projection step}
Due to the impact of the projection step on the stability of the learning, we consider simplifying it and consolidating the environment dynamics. Specifically, we have
\begin{itemize}
    \item A single battery with identical charge/discharge efficiencies
    \item A single renewable generator that cannot be curtailed
    \item No penalties for violating long-term commitment (only price arbitrage)
\end{itemize}
We therefore consider the following state:
\[
x_t = [x^\mathrm{r}_t, \pi^\mathrm{RT}_t, \pi^\mathrm{LT}_t, x^\mathrm{SOC}_t]
\]
and action 
\[
u_t = [u^\mathrm{b}_t, u^\mathrm{LT}_t]
\]
The state space constraints are
\[
0 \leq x^\mathrm{r}_t \leq x^\mathrm{cap} 
\]
\[
\pi^\mathrm{RT}_t, \pi^\mathrm{LT}_t \geq 0
\]
\[
0 \leq x^\mathrm{SOC}_t \leq 1
\]
The action space constraints are
\[
u^\mathrm{min} \leq u^\mathrm{b}_t \leq u^\mathrm{max}
\]
\[
u^\mathrm{LT}_t \geq 0
\]

We can use a hyperbolic tangent to squeeze close or half-closed ranges on the action space. 

Notice that we do not include the real-time commitment in the action space because it is implicitly determined by the other variables:
\[
u^\mathrm{RT}_t = x^\mathrm{r}_t + u^\mathrm{b}_t - u^\mathrm{LT}_t
\]
How do we enforce $u^\mathrm{RT}_t\geq 0$? This is problematic because we need to use the current state to constrain the action. 

In addition, the current SOC depends on the previous SOC which enforces additional constraints on $u^\mathrm{b}_t$.

\newpage
\appendix
\textbf{\huge Appendix}
\section{Offline problem formulation}

Our system model has the following components at a single node:
\begin{itemize}
    \item $B$ batteries
    \item $R$ renewable generators
    \item $N$ conventional dispatchable generators
    \item Exogenous exchange price $\piex$. Settlement is real-time unit pricing.
    \item Pre-existing long-term commitment trajectory
    \item We assume that the frequency of the exchange market is the LCD for all other processes. Dispatch decisions (i.e., charge/discharge, curtailment) are made at the frequency of the exchange market
    \item We care to optimize over a time horizon $t=1,\ldots, T$
    \item Inverter model for renewables?
    \item Do we formulate the problem via state and action spaces?
    \item It may sometimes be advantageous to curtail, for example when the $\piex$ is negative. Question: would it then just be advantageous to store energy in the battery?
    
\end{itemize}
We need to define
\begin{enumerate}
    \item Battery model (system of equations)
    \item Penalty structure for long term commitment
    \item Overall cost function
    \item Representation of uncertain parameters
\end{enumerate}

\begin{subequations}
\label{opt:offline}
\begin{alignat}{2}
    \underset{x_t^\mathrm{ex}, x_t^\mathrm{LT}, x_{r,t}, x_{i,t}, e_{b,t}^\pm, s_{b,t}}{\max}
         \quad &\sum_{t=0}^T f_t^\mathrm{ex}(x_t^\mathrm{ex}; \piex_t) + f_t^\mathrm{LT}(x_t^\mathrm{LT}; \pilt_t,y_t ) -\sum_r f^\mathrm{curtail}_t(x_{r,t}, \overline{x}_{r,t})  \label{eq:cost} \\
    \text{s.t.} \quad & x_t^\mathrm{ex} + x_t^\mathrm{LT} = \sum_{r=1}^R x_{r,t} + \sum_{b=1}^B (e_{b,t}^- - e_{b,t}^+) - \sum_{i=1}^N x_{i,t} \quad &&\forall t\\
    &0 \leq x_t^\mathrm{LT} \leq y_t \quad &&\forall t\\
    &0 \leq x_t^\mathrm{ex} \quad &&\forall t\\
    &\underline{x}_{r,t} \leq x_{r,t} \leq \overline{x}_{r,t} \quad &&\forall r, \ \forall t\\
    &\underline{x}_{i,t} \leq x_{i,t} \leq \overline{x}_{i,t} \quad &&\forall i, \ \forall t\\
    &\underline{e}^+_{b,t} \leq e^+_{b,t} \leq \overline{e}^+_{b,t} \quad &&\forall b, \ \forall t\\
    &\underline{e}^-_{b,t} \leq e^-_{b,t} \leq \overline{e}^-_{b,t} \quad &&\forall b, \ \forall t\\
    &e_{b,t}^+ e_{b,t}^-=0 \quad &&\forall b, \ \forall t\\
    & \underline{s}_{b,t} \leq s_{b,t} \leq \overline{s}_{b,t} &&\forall b, \ \forall t\\
    &s_{b,t+1} = s_{b,t} + \eta_b^+ e_{b,t}^+ - \eta_b^- e_{b,t}^- \quad &&\forall b, \ \forall t
\end{alignat}
\end{subequations}

\textbf{Remarks:}
\begin{enumerate}
    \item The stochastic parameters are $\piex_t$ and  $\overline{x}_{r,t}$. The remaining parameters are given. 
    \item Constraint (1b) is power balance.
    \item The exchange market has a reward function of the form
    \begin{equation}\label{eqn:exchange}
        f^\mathrm{ex}(z; \pi) = \pi z 
    \end{equation}
    where $z \in \R$ and $\pi \in \R_+$. There is no cost associated with generating.
    \item The long-term market has a reward function of the form $f^\mathrm{LT}: \{z|z \leq \overline{z}\} \rightarrow \R$
    \begin{equation}\label{eqn:lt}
        f^\mathrm{LT}(z; \pi, \overline{z}) = \begin{cases}
        \pi z, \quad & z = \overline{z}\\
        g(z), \quad & z < \overline{z}
        \end{cases}
    \end{equation}
    where $g(z)$ is some monotonically increasing function; e.g., constant, linear, stepwise. Question: is $f^\mathrm{LT}$ convex? Like an inverted RElu or some discontinuous function.
    \item An example of the curtailment penalty function is
    \begin{equation}
        f_t^\mathrm{curtail}(z,\overline{z}) = a(z-\overline{z})^2 
    \end{equation}
\end{enumerate}

\section{MPC Formulation}

\begin{itemize}
    \item $H$ be the horizon length for the MPC
    problem.
    \item The state is the level of generation/demand for all generators and ESRs, as well at the SOC for the ESRs. The action will be changes in the state variables. We therefore assume some initial value for the state variables. 
    \item The state $X_t$ is comprised of the physical controllable state $R_t$ and the information state $I_t$ [cite Powell part 2]
    \begin{align*}
        R_t &= (x_{b,t}, x_{r,t}, x_{i,t}, x_t^\mathrm{ex}, x_t^\mathrm{LT})\\
        I_t &= (\pilt, y_t, q_t)
    \end{align*}
    where $q_t$ are the forecasts computed at time $t$. Then
    \begin{equation}
        X_t = (R_t, I_t)
    \end{equation}
    The definition of state is ambiguous... It is unclear how to distinguish certain information $I_t$ from uncertain (exogenous) information $W_t$. For example, we assume $x_{i,t}$ to be known over the whole horizon. But we somehow need to reflect them in our state update.
    \item The action (aka decision) is 
    \begin{equation}
    U_t = (u_{b,t}^\pm, u_{r,t}, u_{i,t}, u_t^\mathrm{ex}, u_t^\mathrm{LT})
    \end{equation}
    Each action must have a corresponding partner in $R_t$. 
    \item Exogenous information is 
    \begin{equation}
        W_t = (\piex, \overline{x}_{r,t})
    \end{equation}
    Exogenous information is information that becomes \textit{first} known between $t-1$ and $t$. This distinguishes $W_t$ from $I_t$ in the state. $W_t$ is assumed to be uncertain. 
    \item The cost function is
    \begin{align*}
    C_t(X_t,U_t, W_t) &= f_t^\mathrm{ex}(x_t^\mathrm{ex}; \piex_t) + f_t^\mathrm{LT}(x_t^\mathrm{LT}; \pilt_t,y_t ) -\sum_r f^\mathrm{curtail}_t(x_{r,t}, \overline{x}_{r,t}) 
    \end{align*}
    
\end{itemize}  

\begin{subequations}
\label{opt:mpc}
\begin{alignat}{2}
    \underset{u_{r,t}, u_{i,t}, e_{b,t}^\pm,u_t^\mathrm{ex}, u_t^\mathrm{LT}}{\max}
         \quad &  \sum_{t=t'}^{t'+H} f_t^\mathrm{ex}(x_t^\mathrm{ex}; \piex_t) + f_t^\mathrm{LT}(x_t^\mathrm{LT}; \pilt_t,y_t ) -\sum_r f^\mathrm{curtail}_t(x_{r,t}, \overline{x}_{r,t})  \\
    \text{s.t.} \quad & x_{r,t+1} = x_{r,t} +  u_{r,t} \quad && \forall r, \forall t\\
    & x_{i,t+1} = x_{i,t} +  u_{i,t} \quad && \forall r, \forall i, \forall t\\
    &x^\mathrm{ex}_{t+1} = x^\mathrm{ex}_{t} + u^\mathrm{ex}_{t} \quad &&  \forall t\\
    &x^\mathrm{LT}_{t+1} = x^\mathrm{LT}_{t} + u^\mathrm{LT}_{t} \quad &&  \forall t\\
    &x_{b,t+1} = x_{b,t} + \eta_b^+ u_{b,t}^+ - \eta_b^- u_{b,t}^- \quad &&\forall b, \ \forall t\\
    &u_t^\mathrm{ex} + u_t^\mathrm{LT} = \sum_{r=1}^R u_{r,t} + \sum_{b=1}^B (u_{b,t}^- - u_{b,t}^+) - \sum_{i=1}^N u_{i,t} \quad &&\forall t\\
    &0 \leq x_t^\mathrm{ex} \quad &&\forall t\\
    &0 \leq x_t^\mathrm{LT} \leq y_t \quad &&\forall t\\
    & \underline{x}_{b,t} \leq x_{b,t} \leq \overline{x}_{b,t} &&\forall b, \ \forall t\\
    &\underline{x}_{r,t} \leq x_{r,t} \leq \overline{x}_{r,t} \quad &&\forall r, \ \forall t\\
    &\underline{x}_{i,t} \leq x_{i,t} \leq \overline{x}_{i,t} \quad &&\forall i, \ \forall t\\
    &\underline{u}^+_{b,t} \leq u^+_{b,t} \leq \overline{u}^+_{b,t} \quad &&\forall b, \ \forall t\\
    &\underline{u}^-_{b,t} \leq u^-_{b,t} \leq \overline{u}^-_{b,t} \quad &&\forall b, \ \forall t\\
    &\underline{u}_{i,t} \leq u_{i,t} \leq \overline{u}_{i,t} \quad &&\forall i, \ \forall t\\
    &u_{b,t}^+ u_{b,t}^-=0 \quad &&\forall b, \ \forall t\\
    &x_{r,t'} = x_{r,t'-1} \quad &&\forall r\\
    &x_{i,t'} = x_{i,t'-1} \quad &&\forall i\\
    &x_{b,t'} = x_{b,t'-1} \quad &&\forall b
\end{alignat}
\end{subequations}

\textbf{Aside:} need to specify terminal constraints such that the transition equation doesn't violate any constraints?

The above problem can be expressed in canonical MPC form:

At each time $t'$, solve
\begin{align*}
\max_{\substack{X_{t'+1}, \ldots, X_{t'+H}\\U_{t'}, \ldots, U_{t'+H-1}}} \quad &\sum_{t=t'}^{t' + H} C_t(X_t, U_t, W_t)\\
\text{s.t.} \quad &X_{t'+1} = f_t(X_t,U_t, W_t), \quad \forall t = t',\ldots,t'+H-1\\
&X_t \in \mathcal{X}_t \quad \forall t = t',\ldots,t'+H\\
&U_t \in \mathcal{U}_t \quad \forall t = t',\ldots,t'+H\\
&W_t \in \mathcal{W}_t \quad \forall t = t',\ldots,t'+H\\
&X_{t'} = X_{t'-1}
\end{align*}

\section{Stochastic Optimization Formulation}
An equivalent Markov-style decision problem can  be formulated: 
\begin{align}
    \max_{\pi \in \Pi} \quad &\mathbb{E}^\pi\sum_{t=0}^T C_t(X_t,  U_t^\pi(X_t))
\end{align}
where the state transition dynamics are 
\[
X_{t+1} = f_t(X_t,U_t^\pi(X_t))
\]
The choice of policy $\pi$ is tbd. If we desire a deterministic lookahead policy with forecasts for the uncertain information, then $U_t^\pi(X_t)$ is given by the argmin of the MPC problem from above. 

\textbf{Question:} does one of the other policy classes better suit this problem?

\section{Literature Review}

In this section, we review the existing approaches to battery and smart grid optimization, highlight the leading methods. 

We organize the scope of the approaches taken to tackle the energy scheduling problem with uncertainty into two main categories: (1) stochastic optimization and (2) reinforcement learning. Each of these classes will be refined further.

Lastly, before starting the review, we state the \textbf{benchmark performance metric}: the \textit{ex-post} (aka posterior, in-hindsight, offline, etc.) optimal. This is when the problem is solved for the entire time horizon using all available information and realizations of uncertainty. In other words, the God solution.



\medskip
\bibliographystyle{abbrvnat}
\bibliography{main}

\begin{thebibliography}{27}
\providecommand{\natexlab}[1]{#1}
\providecommand{\url}[1]{\texttt{#1}}
\expandafter\ifx\csname urlstyle\endcsname\relax
  \providecommand{\doi}[1]{doi: #1}\else
  \providecommand{\doi}{doi: \begingroup \urlstyle{rm}\Url}\fi

\bibitem[Arulkumaran et~al.(2017)Arulkumaran, Deisenroth, Brundage, and
  Bharath]{arulkumaran2017deep}
K.~Arulkumaran, M.~P. Deisenroth, M.~Brundage, and A.~A. Bharath.
\newblock Deep reinforcement learning: A brief survey.
\newblock \emph{IEEE Signal Processing Magazine}, 34\penalty0 (6):\penalty0
  26--38, 2017.

\bibitem[Brockman et~al.(2016)Brockman, Cheung, Pettersson, Schneider,
  Schulman, Tang, and Zaremba]{brockman2016openai}
G.~Brockman, V.~Cheung, L.~Pettersson, J.~Schneider, J.~Schulman, J.~Tang, and
  W.~Zaremba.
\newblock Openai gym.
\newblock \emph{arXiv preprint arXiv:1606.01540}, 2016.

\bibitem[Cao et~al.(2020)Cao, Harrold, Fan, Morstyn, Healey, and
  Li]{cao_deep_2020}
J.~Cao, D.~Harrold, Z.~Fan, T.~Morstyn, D.~Healey, and K.~Li.
\newblock Deep {Reinforcement} {Learning}-{Based} {Energy} {Storage}
  {Arbitrage} {With} {Accurate} {Lithium}-{Ion} {Battery} {Degradation}
  {Model}.
\newblock \emph{IEEE Transactions on Smart Grid}, 11\penalty0 (5):\penalty0
  4513--4521, Sept. 2020.
\newblock ISSN 1949-3061.
\newblock \doi{10.1109/TSG.2020.2986333}.

\bibitem[Cervigni and Perekhodtsev(2013)]{cervigni2013wholesale}
G.~Cervigni and D.~Perekhodtsev.
\newblock Wholesale electricity markets.
\newblock In \emph{The Economics of Electricity Markets}, pages 18--66. Edward
  Elgar Publishing, 2013.

\bibitem[Chen et~al.(2021{\natexlab{a}})Chen, Donti, Baker, Kolter, and
  Berges]{chen2021enforcing}
B.~Chen, P.~Donti, K.~Baker, J.~Z. Kolter, and M.~Berges.
\newblock Enforcing policy feasibility constraints through differentiable
  projection for energy optimization.
\newblock \emph{arXiv preprint arXiv:2105.08881}, 2021{\natexlab{a}}.

\bibitem[Chen et~al.(2021{\natexlab{b}})Chen, Qu, Tang, Low, and
  Li]{chen2021reinforcement}
X.~Chen, G.~Qu, Y.~Tang, S.~Low, and N.~Li.
\newblock Reinforcement learning for decision-making and control in power
  systems: Tutorial, review, and vision.
\newblock \emph{arXiv preprint arXiv:2102.01168}, 2021{\natexlab{b}}.

\bibitem[Fabra(2018)]{fabra2018primer}
N.~Fabra.
\newblock A primer on capacity mechanisms.
\newblock \emph{Energy Economics}, 75:\penalty0 323--335, 2018.

\bibitem[Garifi et~al.(2020)Garifi, Baker, Christensen, and
  Touri]{garifi2020convex}
K.~Garifi, K.~Baker, D.~Christensen, and B.~Touri.
\newblock Convex relaxation of grid-connected energy storage system models with
  complementarity constraints in dc opf.
\newblock \emph{IEEE Transactions on Smart Grid}, 11\penalty0 (5):\penalty0
  4070--4079, 2020.

\bibitem[Helman et~al.(2009)Helman, Singh, and Sotkiewicz]{helman2009rtos}
U.~Helman, H.~Singh, and P.~Sotkiewicz.
\newblock Rtos, regional electricity markets, and climate policy.
\newblock \emph{Generating Electricity in a Carbon-Constrained World}, pages
  527--563, 2009.

\bibitem[Hogan(2016)]{hogan2016virtual}
W.~W. Hogan.
\newblock Virtual bidding and electricity market design.
\newblock \emph{The Electricity Journal}, 29\penalty0 (5):\penalty0 33--47,
  2016.

\bibitem[Kumar et~al.(2021)Kumar, Chandra, Bansal, Kalyanaraman, Ganu, and
  Grant]{kumar2021micro}
P.~Kumar, R.~Chandra, C.~Bansal, S.~Kalyanaraman, T.~Ganu, and M.~Grant.
\newblock Micro-climate prediction-multi scale encoder-decoder based deep
  learning framework.
\newblock In \emph{Proceedings of the 27th ACM SIGKDD Conference on Knowledge
  Discovery \& Data Mining}, pages 3128--3138, 2021.

\bibitem[Litvinov et~al.(2019)Litvinov, Zhao, and
  Zheng]{litvinov2019electricity}
E.~Litvinov, F.~Zhao, and T.~Zheng.
\newblock Electricity markets in the united states: Power industry
  restructuring processes for the present and future.
\newblock \emph{IEEE Power and Energy Magazine}, 17\penalty0 (1):\penalty0
  32--42, 2019.

\bibitem[Mbuwir et~al.(2017)Mbuwir, Ruelens, Spiessens, and
  Deconinck]{mbuwir2017battery}
B.~V. Mbuwir, F.~Ruelens, F.~Spiessens, and G.~Deconinck.
\newblock Battery energy management in a microgrid using batch reinforcement
  learning.
\newblock \emph{Energies}, 10\penalty0 (11):\penalty0 1846, 2017.

\bibitem[Navid and Rosenwald(2012)]{navid2012market}
N.~Navid and G.~Rosenwald.
\newblock Market solutions for managing ramp flexibility with high penetration
  of renewable resource.
\newblock \emph{IEEE Transactions on Sustainable Energy}, 3\penalty0
  (4):\penalty0 784--790, 2012.

\bibitem[Nazir and Almassalkhi(2021)]{nazir2021guaranteeing}
N.~Nazir and M.~Almassalkhi.
\newblock Guaranteeing a physically realizable battery dispatch without
  charge-discharge complementarity constraints.
\newblock \emph{IEEE Transactions on Smart Grid}, 2021.

\bibitem[Oh and Wang(2020)]{oh_reinforcement-learning-based_2020}
E.~Oh and H.~Wang.
\newblock Reinforcement-{Learning}-{Based} {Energy} {Storage} {System}
  {Operation} {Strategies} to {Manage} {Wind} {Power} {Forecast} {Uncertainty}.
\newblock \emph{IEEE Access}, 8:\penalty0 20965--20976, 2020.
\newblock ISSN 2169-3536.
\newblock \doi{10.1109/ACCESS.2020.2968841}.

\bibitem[Powell(2014)]{powell2014clearing}
W.~B. Powell.
\newblock Clearing the jungle of stochastic optimization.
\newblock In \emph{Bridging data and decisions}, pages 109--137. Informs, 2014.

\bibitem[Powell(2021)]{powell2021reinforcement}
W.~B. Powell.
\newblock Reinforcement learning and stochastic optimization, 2021.

\bibitem[Powell and Meisel(2015)]{powell2015tutorial}
W.~B. Powell and S.~Meisel.
\newblock Tutorial on stochastic optimization in energy—part ii: An energy
  storage illustration.
\newblock \emph{IEEE Transactions on Power Systems}, 31\penalty0 (2):\penalty0
  1468--1475, 2015.

\bibitem[Schulman et~al.(2017)Schulman, Wolski, Dhariwal, Radford, and
  Klimov]{schulman2017proximal}
J.~Schulman, F.~Wolski, P.~Dhariwal, A.~Radford, and O.~Klimov.
\newblock Proximal policy optimization algorithms.
\newblock \emph{arXiv preprint arXiv:1707.06347}, 2017.

\bibitem[Shang et~al.(2020)Shang, Wu, Guo, Ma, Sheng, Lv, and
  Fu]{shang_stochastic_2020}
Y.~Shang, W.~Wu, J.~Guo, Z.~Ma, W.~Sheng, Z.~Lv, and C.~Fu.
\newblock Stochastic dispatch of energy storage in microgrids: {An} augmented
  reinforcement learning approach.
\newblock \emph{Applied Energy}, 261:\penalty0 114423, Mar. 2020.
\newblock ISSN 0306-2619.
\newblock \doi{10.1016/j.apenergy.2019.114423}.
\newblock URL
  \url{https://www.sciencedirect.com/science/article/pii/S0306261919321105}.

\bibitem[Tang et~al.(2022)Tang, Liu, and Zeng]{tang2022multi}
Z.~Tang, J.~Liu, and P.~Zeng.
\newblock A multi-timescale operation model for hybrid energy storage system in
  electricity markets.
\newblock \emph{International Journal of Electrical Power \& Energy Systems},
  138:\penalty0 107907, 2022.

\bibitem[Wang and Zhang(2018)]{wang_energy_2018}
H.~Wang and B.~Zhang.
\newblock Energy storage arbitrage in real-time markets via reinforcement
  learning.
\newblock In \emph{2018 {IEEE} {Power} and {Energy} {Society} {General}
  {Meeting}, {PESGM} 2018}, page 8586321. IEEE, Institute of Electrical and
  Electronics Engineers, 2018.
\newblock \doi{10.1109/PESGM.2018.8586321}.
\newblock URL
  \url{https://research.monash.edu/en/publications/energy-storage-arbitrage-in-real-time-markets-via-reinforcement-l}.

\bibitem[Yang et~al.(2020{\natexlab{a}})Yang, Yang, Wang, Du, and
  Yu]{yang_deep_2020}
J.~J. Yang, M.~Yang, M.~X. Wang, P.~J. Du, and Y.~X. Yu.
\newblock A deep reinforcement learning method for managing wind farm
  uncertainties through energy storage system control and external reserve
  purchasing.
\newblock \emph{International Journal of Electrical Power \& Energy Systems},
  119:\penalty0 105928, July 2020{\natexlab{a}}.
\newblock ISSN 0142-0615.
\newblock \doi{10.1016/j.ijepes.2020.105928}.
\newblock URL
  \url{https://www.sciencedirect.com/science/article/pii/S0142061519312505}.

\bibitem[Yang et~al.(2020{\natexlab{b}})Yang, Zhao, Li, and
  Zomaya]{yang2020reinforcement}
T.~Yang, L.~Zhao, W.~Li, and A.~Y. Zomaya.
\newblock Reinforcement learning in sustainable energy and electric systems: A
  survey.
\newblock \emph{Annual Reviews in Control}, 49:\penalty0 145--163,
  2020{\natexlab{b}}.

\bibitem[Zhang et~al.(2019)Zhang, Zhang, and Qiu]{zhang2019deep}
Z.~Zhang, D.~Zhang, and R.~C. Qiu.
\newblock Deep reinforcement learning for power system applications: An
  overview.
\newblock \emph{CSEE Journal of Power and Energy Systems}, 6\penalty0
  (1):\penalty0 213--225, 2019.

\bibitem[Zheng et~al.(2021)Zheng, Shi, Ratliff, and Zhang]{zheng2021safe}
L.~Zheng, Y.~Shi, L.~J. Ratliff, and B.~Zhang.
\newblock Safe reinforcement learning of control-affine systems with vertex
  networks.
\newblock In \emph{Learning for Dynamics and Control}, pages 336--347. PMLR,
  2021.

\end{thebibliography}

\end{document}